\documentclass{article}

\usepackage{microtype}
\usepackage{graphicx}
\usepackage{subfigure}
\usepackage{booktabs} 

\newcommand{\method}[0]{\text{RARE}}

\newcommand{\z}[0]{\mathbf{z}}
\newcommand{\s}[0]{\mathbf{s}}
\newcommand{\ac}[0]{\mathbf{a}}
\newcommand{\cbold}[0]{\mathbf{c}}

\usepackage{hyperref}

\usepackage[accepted]{icml2024}

\usepackage{amsmath}
\usepackage{amssymb}
\usepackage{mathtools}
\usepackage{amsthm}

\usepackage[capitalize,noabbrev]{cleveref}

\theoremstyle{plain}

\theoremstyle{definition}

\theoremstyle{remark}

\usepackage[textsize=tiny]{todonotes}

\icmltitlerunning{Leveraging Human Revisions for Improving Text-to-Layout Models}

\begin{document}

\twocolumn[
\icmltitle{Leveraging Human Revisions for Improving Text-to-Layout Models}



\icmlsetsymbol{equal}{*}

\begin{icmlauthorlist}
\icmlauthor{Amber Xie}{comp}
\icmlauthor{Chin-Yi Cheng}{comp}
\icmlauthor{Forrest Huang}{comp}
\icmlauthor{Yang Li}{comp}
\end{icmlauthorlist}

\icmlaffiliation{comp}{Google Research, Mountain View, USA}

\icmlcorrespondingauthor{Chin-Yi Cheng}{cchinyi@google.com}
\icmlcorrespondingauthor{Yang Li}{liyang@google.com}

\icmlkeywords{Machine Learning, ICML}

\vskip 0.3in
]



\printAffiliationsAndNotice{} 

\begin{abstract}
Learning from human feedback has shown success in aligning large, pretrained models with human values. Prior works have mostly focused on learning from high-level labels, such as preferences between pairs of model outputs. On the other hand, many domains could benefit from more involved, detailed feedback, such as revisions, explanations, and reasoning of human users. Our work proposes using nuanced feedback through the form of human revisions for stronger alignment. In this paper, we ask expert designers to fix layouts generated from a generative layout model that is pretrained on a large-scale dataset of mobile screens. Then, we train a reward model based on how human designers revise these generated layouts. With the learned reward model, we optimize our model with reinforcement learning from human feedback (RLHF). Our method, Revision-Aware Reward Models ($\method$), allows a generative text-to-layout model to produce more modern, designer-aligned layouts, showing the potential for utilizing human revisions and stronger forms of feedback in improving generative models.
\end{abstract}

\section{Introduction}
Large, pretrained models have shown impressive results in many domains, including natural language and text-to-image generation. However, because these models are typically trained on large-scale, unfiltered data, they may not be aligned with human values. To ensure positive usage of these models. it is important to address the issue of unsafe, inaccurate, or outdated generations.

A developing area of study attempts to address the misalignment problem by learning from human feedback. Many of these works have primarily focused on using high-level labels of human feedback, e.g., preferences between pairs of model outputs such as images \cite{lee2023aligning} or languages \cite{liu2023chain}. However, in the real world, we often learn better when relying on detailed corrections, explanations, and reasoning. We hypothesize that learning from human revisions is more effective for a model to adapt to produce human preferred results. Compared to preferences or language feedback, which has been used in prior works, revisions are a stronger type of feedback that indicate human preferences on the end results and provide nuances in how to align model outputs with human expectations. 

In this paper, we investigate this approach in the domain of text-to-layout generation, in which a model is trained to generate a layout given a text input. 
To do so, we first ask professional designers to improve layouts generated from PLay~\citep{cheng2023play}, a generative layout model that is trained on a large-scale dataset of mobile screens that reflects earlier generations of Android UIs. To modernize and improve layouts, designers perform revisions in Figma, a popular design tool. Our plugin records detailed, step-by-step edits that designers perform in revising a layout towards their satisfaction. Based on these revision sequences, we train a reward model, which is then used to optimize the generative model, using reinforcement learning from human feedback (RLHF). 

Based on our experiments, our method, Revision-Aware Reward Models ($\method$), outperforms the baseline method that directly fine-tunes a generative model with a supervised approach. Our experiments also showed that reward signals that are designed based on revision sequences lead to more desirable outcomes than using preferences alone. By analyzing the outputs acquired by our method, we found our method leads to more modern, designer-aligned layouts, even though the base model was trained on a dataset with old-fashioned Android designs. This shows the potential for utilizing human revisions and stronger forms of feedback in improving text-to-layout models.

\textbf{Contributions} We highlight our contributions as follows:
\begin{itemize}
    \item We collect a high-quality, detailed dataset  from experienced designers, producing over 2,000 sequences of revisions from the generated layout to the human-revised layout.
    \item We propose a method, $\method$, for learning from human revisions of layout design, and examine two variants of $\method$, including Keystroke and Chamfer.
    \item Our experiments and analysis show that reward models that use human improvement-specific data can be used in RLHF to effectively finetune the base text-to-layout model for qualitative and quantitative improvements.
\end{itemize}

\begin{figure*}[th!]
    \centering
    \includegraphics[width=\textwidth]{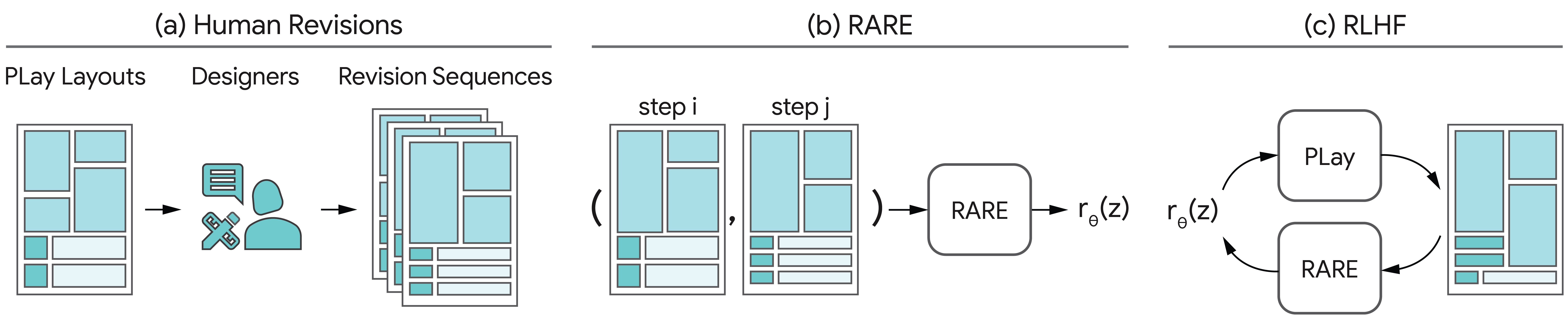}
    \caption{\textbf{$\method$ Method Overview} Our method consists of three parts: (a) Collecting human revision sequences, (b) Training our reward model on sequence data, and (c) Using the reward model in an RLHF framework.}
    \label{fig:method}
\end{figure*}

\section{Related Work}
Human priors have been used to guide language model alignment \citep{liu2023chain, ouyang2022training}, text-to-image models \citep{lee2023aligning, fan2023dpok, black2023training}, and behaviors \citep{hejna2022fewshot, lee2021pebble}. These human priors can be represented through (1) Curating high quality, human-aligned datasets for finetuning \citep{ouyang2022training}; (2) Extracting rewards from pretrained foundation models like Stable Diffusion \citep{rombach2022highresolution, jain2022vectorfusion} or R3M \citep{nair2022r3m, adeniji2023language};  (3) Explicitly learning reward functions from human feedback \citep{xu2023imagereward, lee2023aligning, bai2022training, stiennon2022learning}. Our work utilizes human feedback for alignment.

\textbf{Types of Human Feedback} The most informative types of human feedback is still an open area of research. Prior works in generative models have primarily used binary human preferences \citep{bai2022training, liu2023chain, lee2023aligning}, or scalar ratings \citep{stiennon2022learning}. To our knowledge, there are no works that actively leverage human revisions for generative models, which is a stronger and more involved form of human feedback we propose.

Notably, correctional feedback has been used in robotics applications \citep{li2021learning, losey2021physical, 9473611, ross2011reduction}. However, these robotics-focused works focus on improving a trajectory or a sequence of actions, which is a multi-step bandit problem. Our work focuses on improving generative samples, which is a one-step contextual bandit problem. 

\textbf{Learning from Human Feedback} Given human-annotated data, a popular approach is reinforcement learning from human feedback (RLHF). RLHF consists of a two-stage process: (1) Training a reward model on human feedback and (2) Optimizing a reinforcement learning objective. This has shown success in language modelling \citep{casper2023open}, text-to-image models \citep{lee2023aligning, black2023training}, and more.

Reinforcement learning-free methods can also learn from human feedback. A traditional method is supervised finetuning on the curated dataset \citep{ouyang2022training}. Recent papers have proposed new supervised objectives, such as Chain of Hindsight \citep{liu2023chain} and Direct Policy Optimization \citep{rafailov2023direct}, to align large language models with human feedback.

\textbf{Layout Generation} For generation tasks such as layout design, layouts are saved in a vector graphic format, so that designers can easily edit and use them for downstream purposes. This modality is amenable  for collecting human revisions. Recent studies on layout generation use sequential models like Transformers \citep{transformer} to output the layout elements as sequences \citep{gupta2021layouttransformer, vtn, blt, kikuchi2021constrained}. LayoutDM and PLay \citep{inoue2023layoutdm, cheng2023play} show results in conditional layout generation. We choose PLay as our backbone model based on its flexibility to inject various conditions using latent diffusion \citep{ldm} and classifier-free guidance \citep{cfg}.

\section{Background}
\subsection{Diffusion Models}
Diffusion models are a popular class of generative models that learns a denoising processing from a known Gaussian prior to the data distribution. During training, the diffusion model learns to reverse a single noising step, which reduces to the following training objective:

\begin{equation}
\mathcal{L}_{DDPM} (\phi, \z) = \mathbb{E}_{t, \epsilon} [w(t) || \epsilon_{\phi} (\alpha_t \z + \sigma_t \epsilon) - \epsilon ||^2]
\end{equation}

where $\phi$ are the learned parameters, $\z$ is a real data sample, $\epsilon$ is the added noise, $t$ is a scalar time step that schedules noise values $\alpha_t, \sigma_t$.

To sample from the diffusion model, the diffusion model iteratively denoises the initial sample $\z_T$ from the known prior $N(0, 1)$, where $T$ is the number of denoising steps, and $x_0$ is the final sampled data point. Denoising steps are modelled as Gaussian distributions:

\begin{equation}
    p_\phi (\z_{t-1} | \z_t) = N(\z_{t-1} | \epsilon_\phi(\z_t, t), \sigma^2_t I)
\end{equation} 

Diffusion models can be conditioned on additional context $\cbold$, whether it is text \citep{rombach2022highresolution}, guidelines \citep{cheng2023play}, or more. Furthermore, to reduce computational costs, diffusion models are often trained in latent space.

\subsection{Generative Layout Models}
Layout designs are used by engineers and designers to produce vectorized arrangements and models. There are several commonly used datasets for layout modeling, including PublayNet \citep{zhong2019publaynet}, CLAY \citep{li2021learning}, and RICO-Semantic \citep{https://doi.org/10.48550/arxiv.2210.02663}. In this paper, we focus on UI layouts, which consist of a collection of UI elements. Each UI element has a class type such as Button or Checkbox, and a bounding box that specifies the position and size of the element. In particular, we look at the task of generating UI Layouts based on text input, which is useful for assisting UI designers to create design mockups.

In this work, we train PLay~\citep{cheng2023play} to generate a UI layout based on a text input, such as ``a login screen". We condition on text to provide context for the layout design, aiding designers to generate meaningful revisions.

\begin{figure*}
    \centering
    \includegraphics[width=\textwidth]{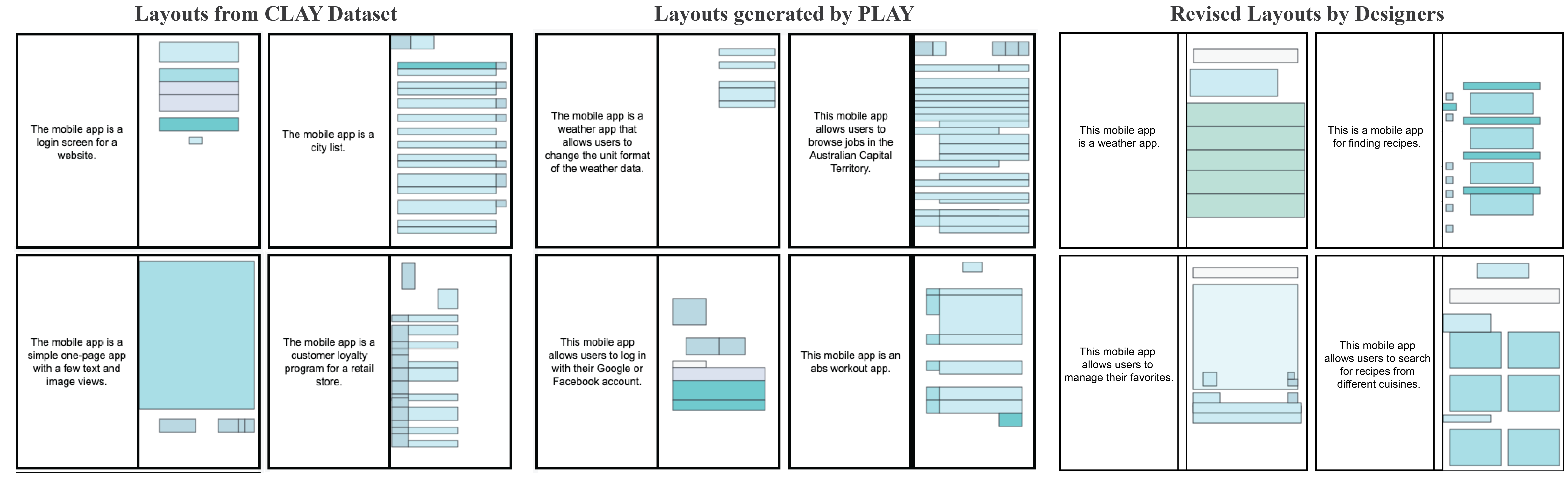}
    \caption{\textbf{Layouts Visualizations} We use CLAY (left) as our pretraining dataset. PLay (middle) is a generative, text-conditioned layout model. We ask designers to edit layouts generated by PLay, leading to more modern, cohesive layouts (right).}
    \label{fig:samples}
\end{figure*}

\section{Method}
\label{section:method}

Our objective is to better align a generative layout model, such as PLay \citep{cheng2023play}, with human revisions. First, we collect a dataset of revisions from experienced designers. Next, we learn a reward model from the human revision dataset. Finally, we use reinforcement learning from human feedback (RLHF) to improve our base model. 

\subsection{Human Revision Data Collection}

\textbf{Figma Plugin} To facilitate data collection, we use Figma\footnote{\url{https://www.figma.com/}}, a design tool commonly used by designers. Our Figma plugin visualizes mock-ups generated by PLay and its corresponding text condition. The plugin uses text boxes and elements from Material 3 design library\footnote{\url{https://m3.material.io/}} to represent the various classes. Our plugin records all the operations performed by the designer as well as corresponding layout states in the process of revising a layout.

\textbf{Human Revisions} We recruit 4 professional designers to revise layouts generated from PLay. Designers are asked to modify the layout to be aesthetic and coherent, and they are provided an instructions and example revisions. For instance, we expect designers to fix misaligned elements or change the format of the layout based on the text description. 

Designers are able to move, scale, and change classes of elements. Designers may add or remove elements as necessary. We conduct the study asynchronously, without time restrictions. After the designer completes the task, we save the sequence of revisions and the final layout.

Our final dataset $\mathcal{D}_{\text{human}}$ consists of revision sequences $\{ \{ b^j_i, l^j_i \}_{j=0}^M,
d_i\}_{i=0}^N$. At each revision step $i$,  $b^j_i$ and $l^j_i$ are the bounding boxes and class labels for the $M$ elements in layout. $d_i$ is the Chamfer Distance of the $i$th layout to the final layout. Finally, $N$ is the number of total revisions the designer makes. 

\subsection{Reward Learning}
\label{headings}
To maximally leverage the revision data, we propose \textbf{Revision-Aware Reward Models ($\method$)}. $\method$ predicts the amount of effort needed to improve the layout. Unlike preference or binary ratings, $\method$ produces a scalar value roughly corresponding to the quality of the data point: low revision effort implies a strong layout, and high revision effort implies a poor layout.

In practice, we define the effort as the \textbf{Chamfer Distance} between two layouts, where a low Chamfer Distance should imply a layout close to final, revised layout, and a high Chamfer Distance implies a layout far from the final layout.  

At each revision step of our revision sequence, we calculate the Chamfer Distance $d$ of the intermediate layout to the final layout, and the latent embedding $z$ of the intermediate layout, obtained by using PLay's first stage latent autoencoder. From this, we construct a supervised learning problem for our reward model $r_\theta$:
\begin{equation}
\mathcal{L}_\text{RARE} = \mathbb{E}_{(\mathbf{z}, t) \sim \mathcal{D_{\text{human}}}} [(r_\theta(z) - d)^2]
\end{equation}

Under the framework of Revision-Aware Reward Models ($\method$), we may also use different reward models that are extract revision-based rewards. An ablation of this method is the \textbf{Keystroke Model}, which is based on the amount of time needed for the designer to complete the revision of the layout. The Keystroke reward model is trained to predict the time it takes a designer to reach the final revised layout from an intermediate layout. The Keystroke time between two layouts is an alternative way to capture the amount of effort needed to improve the layout, where a high Keystroke time implies a poor layout. We train this reward model under the same supervised objective, replacing the Chamfer Distance $d$ with the Keystroke time.

\subsection{Reinforcement Learning from Human Feedback}

Finally, to align our model with human feedback, we use RLHF. Following \citet{black2023training}, we treat the learned denoising process as a Markov Decision Process, where the environment is represented as the following:
\begin{equation}
\begin{split}
\s_t &\triangleq (\cbold, t, \z_t) \;\;\;\;\;\;\;\;
a_t \triangleq \z_{t-1} \\ 
P(\s_{t+1} | \s_t, \ac_t) &\triangleq (\delta_\cbold, \delta_{t-1},\delta_{\z_{t-1}}) \\ 
\pi(\ac_t | \s_t) &\triangleq p_\phi(\z_{t-1} | \z_t, \cbold) \\
\mathcal{R}(\s_t,\ac_t) &\triangleq 
\begin{cases}
r_\theta(x_0, \cbold) & \text{if t = 0}\\
0 & \text{otherwise} 
\end{cases} \\
\rho_0 &\triangleq (p(\cbold), \delta_T, \mathcal{N}(0, I)) 
\end{split}
\end{equation}

where $c$ is the conditioned text, $\pi$ is our policy, $\rho_0$ is the initial state distribution, $\delta$ is the Dirac delta, $T$ is the number of sampling steps, and $r_{\theta}(x_0, \cbold)$ is our reward $\method$. 

We optimize for the reward by using DDPO with importance sampling. The DDPO algorithm is based on Proximal Policy Optimization \citep{schulman2017proximal}, which clips importance sampling weights to constraint update steps. More about RLHF algorithm is in Appendix~\ref{appendix:rlhf_hp}.

Our method, Revision-Aware Reward Models ($\method$), is simple and can be easily integrated in existing RLHF literature and algorithms. Our main contribution is constructing a more complete dataset, consisting of the entire revision sequence, which allows us to train stronger reward models. Then, by optimizing reward models based on nuanced feedback, reinforcement learning algorithms can more effectively align pretrained models with human values.

\begin{figure*}
    \centering
    \includegraphics[width=0.8\textwidth]{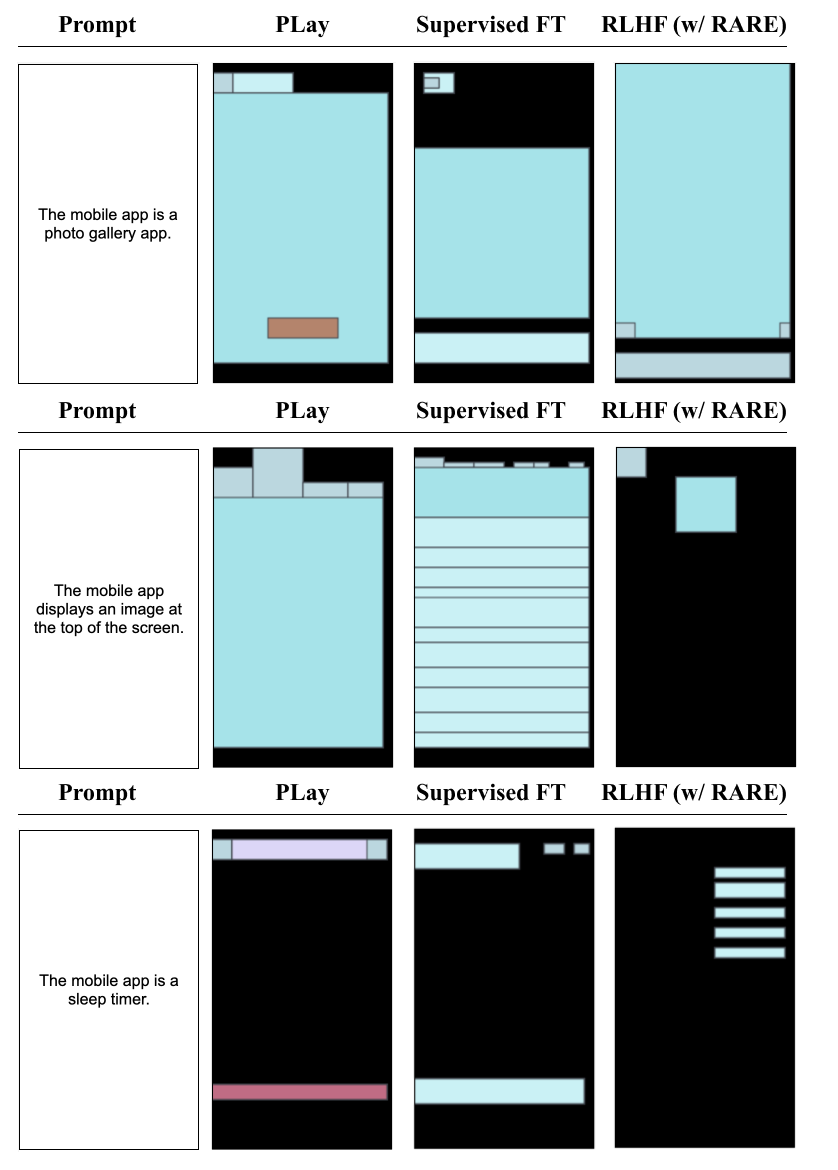}
    \caption{\textbf{Result Comparison} We compare layouts generated by a PLay model, a supervised finetuned model, and a model trained with RLHF with $\method$. In these examples, RLHF w/ $\method$ produces the most cohesive and aligned layouts.}
    \label{fig:play_ft_rare}
\end{figure*}

\section{Experiments}
\label{others}

\subsection{Experimental Setup}

\textbf{Base Model} We use a variation of the PLay model, which is conditioned on text input instead of grid-based guidelines in the original paper. We pretrain the text-conditioned PLay, as our base model, on CLAY~\citep{clay}, a public dataset, with corresponding text labels from CLAY screens generated by PaLM~\citep{chowdhery2022palm}, an LLM. The CLAY dataset is derived from RICO~\citep{rico}, a popular mobile corpus that consists of UI screens of early Android verions. The hyperparameters for training text-conditioned PLay are included in the Appendix.

\begin{figure}[t]
\begin{center}
\includegraphics[width=\linewidth]{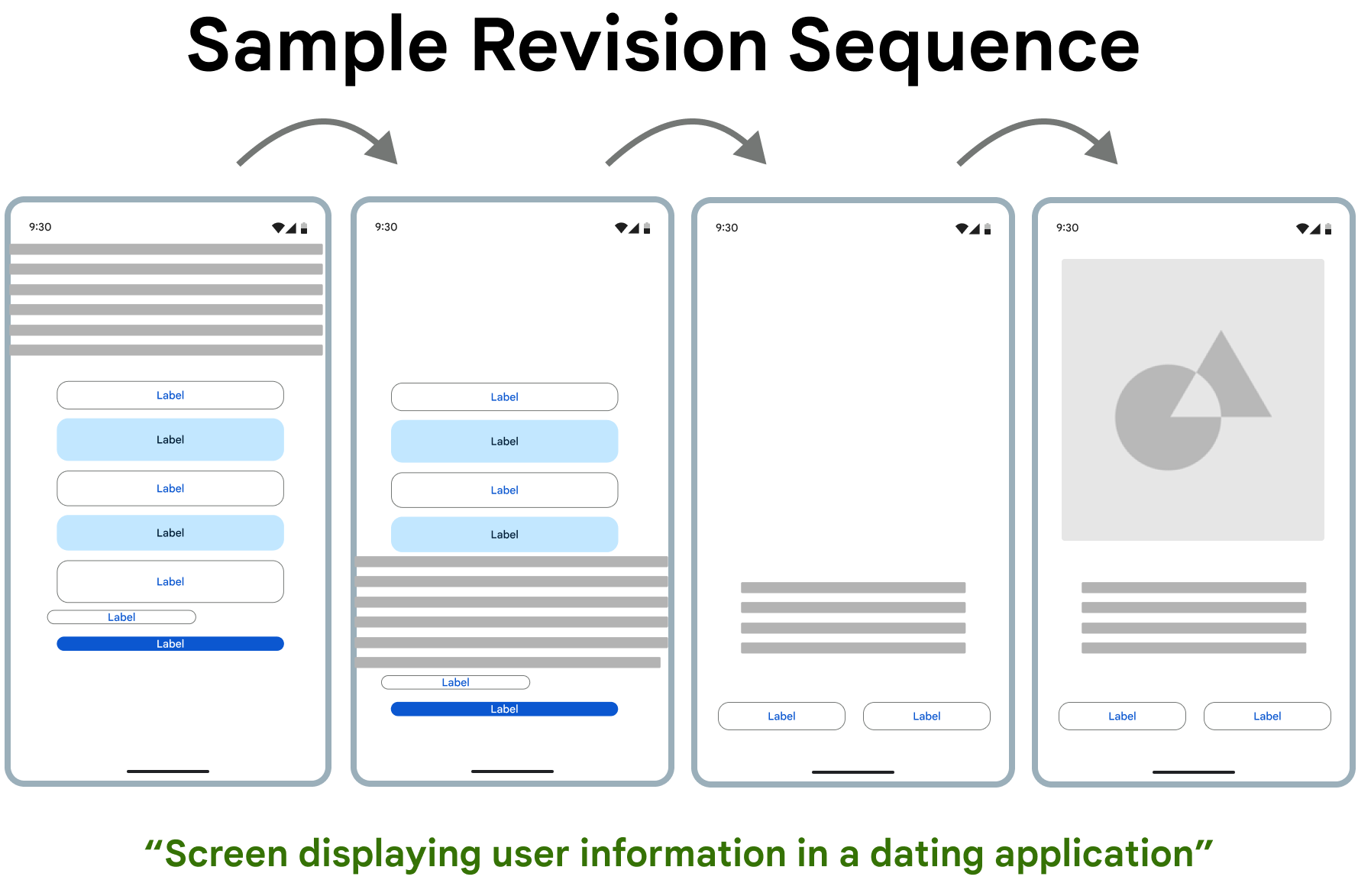}
\end{center}
\caption{\textbf{Figma Plugin} Our Figma plugin renders a PLay layout with the corresponding text description. Designers are asked to revise the layout by adding, modifying, and deleting elements.}
\end{figure}

\begin{figure}
\includegraphics[width=\linewidth]{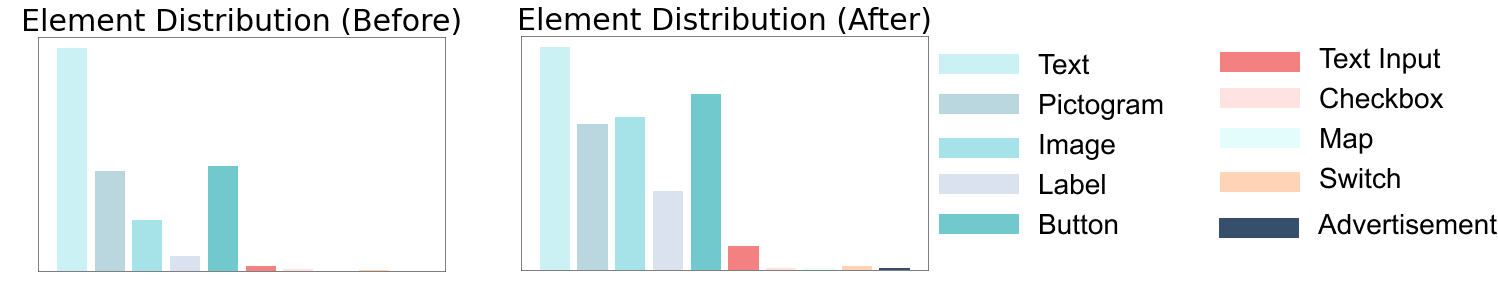}
\label{fig:sfig1}
\caption{\textbf{Element Distribution} The element distribution from PLay samples (left) becomes more diverse after revisions (right). Designers add more elements  (mean number of elements increases from 11.02 to 13.05 after revisions), particularly images and labels.}
\label{fig:elem_distribution}
\end{figure}

\textbf{Dataset} We use a dataset of 836 revised UI Layouts from designers. We extract every tenth layout as part of the revision sequence, to account for redundancy in the logging system. Then, the average revision sequence length is 88.9, with 8,694 unique layouts total. We split the dataset into 645 train and 191 evaluation examples. Statistics on edit time and more are included in Figure~\ref{fig:dataset_stats}, and distribution shifts in element classes are in Figure~\ref{fig:elem_distribution}.

Furthermore, because our revision dataset is limited, we generate a synthetic dataset for pretraining the reward models. We generate intermediate layouts for the final revised layout by randomly revising, adding, and dropping elements. Given pairs, we can calculate Chamfer Distance or Keystroke time to generate a synthetic dataset for pretraining our reward models. More details are in Appendix~\ref{appendix:proceduralpretrain}. After pre-training the reward model on this synthetic dataset, we efficiently finetune our reward model on the revision dataset.

\textbf{Baselines} We evaluate the following methods:
\begin{enumerate}
    \item Supervised Finetuning (SFT): We directly finetune PLay on the final revisions.
    \item Preference Reward + RLHF: We train a Preference reward model on pairs from the revision sequences. We assume that the final revision is the most optimal, so all intermediate revisions are considered negatives.
    \item $\method$ Chamfer Distance + RLHF: We learn the Chamfer Distance from an intermediate layout to the final layout and utilize the negative distance as the reward. The Chamfer Distance is a geometric distance that is revision-aware ($\method$).
    \item $\method$ Keystroke + RLHF: We learn the time difference from an intermediate layout to the final layout and utilize the negative predicted time as the reward. This time-based metric is revision-aware ($\method$).
\end{enumerate}

All rewards are normalized for more efficient RL training. For our RLHF methods, because our objective is to lightly finetune our model, we use a standard DDPM sampler, but we only optimize the DDPO objective on the last 10 timesteps of the denoising diffusion model. We find that further optimization for early steps of the denoising process leads to mode collapse and reward exploitation. For additional hyperparameters, please refer to the Appendix.

\subsection{Quantitative Results}

\begin{table*}
\centering
\begin{tabular}{llll}
\toprule
\multicolumn{2}{c}{} & \textbf{FID Score ($\downarrow$)} & \textbf{DocSim ($\uparrow$)} \\
\midrule
Dataset & CLAY Dataset & 73.4 $\pm$ 4.5 & 0.14 $\pm$ 0.01
 \\
& Revision Dataset & 63.1 $\pm$ 3.0 & 0.16 $\pm$ 0.01 \\
& PLay Samples & 76.6 $\pm$ 0.3 & 0.31 $\pm$ 0.002 \\
\midrule
Finetuning & Supervised FT & 68.9 $\pm$ 0.6
 & 0.26 $\pm$ 0.01 \\
\midrule
RLHF & $\method$ Keystroke  &  70.1 $\pm$ 0.8 & 0.27 $\pm$ 0.03
 \\
& $\method$ Chamfer &  68.8 $\pm$ 0.5 & 0.28 $\pm$ 0.01 \\
& Preference  &  72.4 $\pm$ 2.6 & 0.31 $\pm$ 0.02 \\
\bottomrule
\end{tabular}
\caption{We evaluate the FID Score and DocSim with a held-out batch of human revisions. For calibration, we include the FID scores of dataset samples from CLAY, the Revision Dataset, and PLay Samples. We find that $\method$ Keystroke exceeds Supervised Finetuning in DocSim and matches Supervised Finetuning in FID score.}
\label{tab:fid}
\end{table*}

To evaluate the alignment of our finetuned model with the human revision dataset, in Table~\ref{tab:fid}, we report the FID scores ($\downarrow$) with the final revisions; and the DocSim scores ($\uparrow$) \cite{DBLP:journals/corr/abs-1909-00302}, a measure of similarity across documents. 

First, we calculate the FID scores ($\downarrow$) between samples from existing datasets and a held-out batch of final revised layouts. The FID score between training and validation revised layouts (Revised Dataset) is the lowest, as both layouts are from the same, revised distribution. Both CLAY and PLay layouts have higher FID scores, suggesting there is a noticeable distribution shift between the CLAY dataset and revised layouts. However, we are able to achieve lower FID scores than CLAY and PLay samples, showing that supervised finetuning and RLHF methods are able to improve upon existing datasets and pretrained models.

We find that the DocSim scores ($\uparrow$) for PLay model samples are higher than DocSim scores for the CLay and Revision Datasets. We hypothesize that CLAY dataset layout elements are quite different in size and positioning to Revised Layouts, as there are often cluttered, smaller elements from older Android designs. In addition, we hypothesize that revised layouts from designers may vary greatly in styles and sizes of elements. Although in a more modern, designer-preferred format, the modes are more extreme per layout, such as very sparse layouts or very detailed layouts, thus possible leading to a lower DocSim score. We include samples from these different datasets in Figure~\ref{fig:samples}.

Through supervised finetuning and RLHF methods, we are able to improve samples beyond PLay generated samples and even the original training dataset, CLAY. We find that $\method$ Chamfer produces the lowest FID score, slightly outperforming supervised finetuning. In DocSim scores, $\method$ Chamfer and all other RLHF methods outperform supervised finetuning, suggesting that RLHF methods are able to better align pretrained models with human values.

Among RLHF methods, we find that $\method$ methods, which take advantage of the entire revision sequence, is able to outperform the Preference reward model. Even though the Preference reward model is trained on the entire revision sequence, the dataset is processed into pairs of preferred layouts, without weighing each intermediate layout differently based on where it is in the revision sequence. We hypothesize that the strength of our method is in utilizing the revision information in our reward models.

Finally, we find that using the $\method$ Keystroke is far less effective than $\method$ Chamfer Distance. This result is interesting, because both quantify the human effort needed to revise the layout. We hypothesize that the Keystroke time is much harder to learn, especially on a limited dataset. The variance in the amount of time designers take, further explored in Section~\ref{sec:datasetanalysis}, may make Keystroke time difficult to predict. In addition, designers work at different paces, making the Keystroke prediction more difficult.

\begin{figure*}
    \centering
    \includegraphics[width=0.8\linewidth]{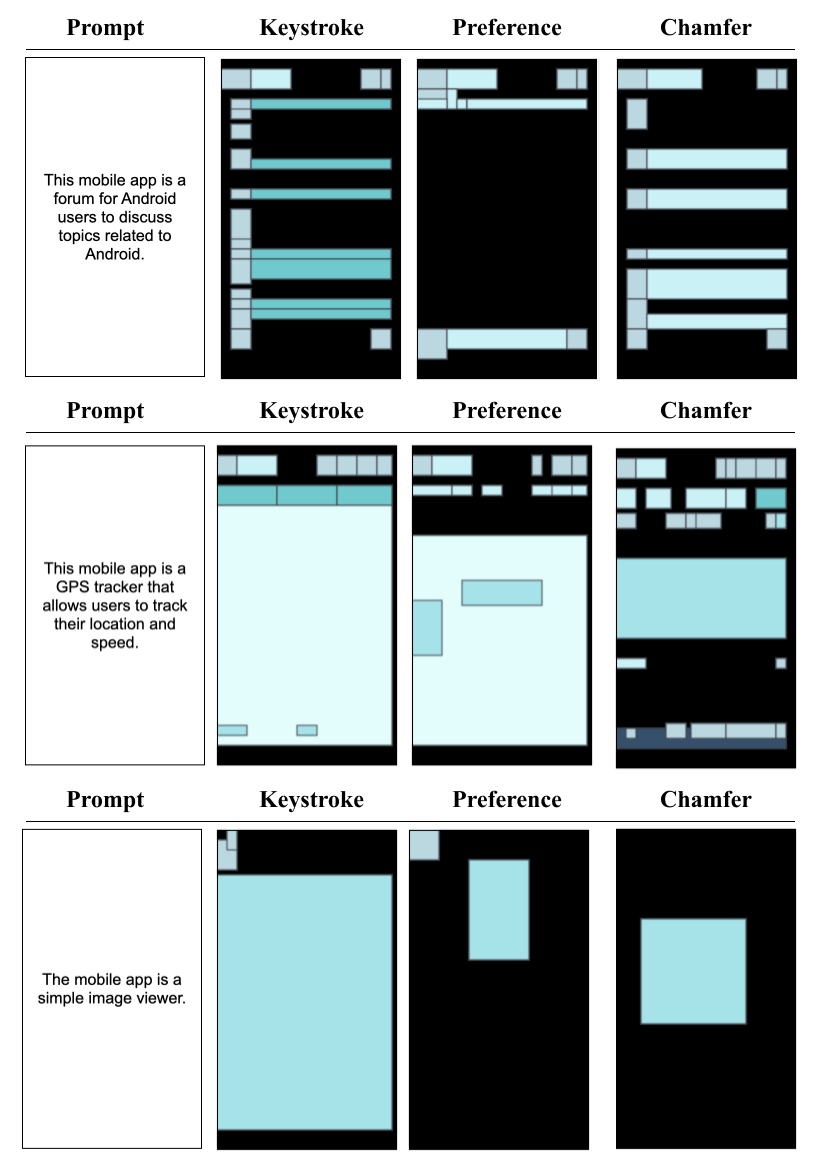}
    \caption{\textbf{RLHF with Different Reward Models} We compare the effect of using different reward models. We find that RARE Keystroke and RARE Chamfer lead to more consistent and coherent samples. For instance, in the first row, the RARE Chamfer and RARE Keystroke samples may resemble forum discussions more. In the second example, RARE Chamfer and Keystroke samples are well-aligned. In the last example, we find that RARE Keystroke generates a large image, which is typically unseen in the pretraining data, suggesting successful alignment with human revisions.}
    \label{fig:reward_ablation}
\end{figure*}

\subsection{Qualitative Comparisons}
We show qualitative examples of how $\method$ may better align layouts with user preferences. In Figure~\ref{fig:play_ft_rare}, we compare layouts from PLay (the base model), PLay + Supervised Finetuning, and PLay + RLHF w/ $\method$. We find that samples from the Supervised FT model differ greatly from the base model and are inconsistent in quality. However, RLHF w/ RARE stays close to the base model samples, slightly improving it based on the learned design principles. This supports our hypothesis that RARE provides informative feedback and better alignment.

Next, we compare how different reward models trained on the revision sequences affect the quality of the generated outputs. In Figure~\ref{fig:reward_ablation}, we compare $\method$ with Keystroke, Preference, and Chamfer Distance reward models. Although RLHF with Preference reward models achieves a similar FID score to $\method$, quantitatively, we notice qualitative differences in alignment and overall layout, described further in detail in Figure~\ref{fig:reward_ablation}.

\subsection{Dataset Analysis}
\label{sec:datasetanalysis}
We provide additional insight to how designers are revising layouts in Figures~\ref{fig:dataset_stats}. Designers take a median time of 503.4 seconds per full revision. The mean number of elements before is 10.4 (standard deviation 7.9), and the mean number of elements after revision is 15.8 (standard deviation 9.0).

In addition, we plot the element class distributions from before and after revisions in Figure ~\ref{fig:elem_distribution}. From the revision data and designer feedback, we find that the base PLay generations are not optimal for most use cases. The shift in element class distributions and the amount of time spent revising the layouts suggests that many elements are misplaced or of the wrong class. This may be partially due to the fact that the base model PLay was trained on an older dataset of UI screens, motivating the need for human alignment.

\section{Conclusions}
In this work, we present a method for leveraging detailed human feedback through the form of revision sequences. We ask designers to revise layouts generated a text-conditioned generative layout model. Our method, $\method$ uses the revision data to train revision-aware reward models. The $\method$ framework is easily incorporated into existing RLHF algorithms and successfully improves the pretrained model to produce more modern designer-aligned layouts. 

In addition, we provide thorough results and analysis on ablations to $\method$. For example, we show results on different reward functions trained on the revision sequences. Our strongest reward model is $\method$ Chamfer Distance model, which predicts the geometric distance between an intermediate and the final revision, which provides a stronger reward signal that previous reward models based on human feedback. $\method$ has strong quantitative and qualitative results, leading to layouts that are well-aligned, cohesive, and more aligned with human preferences.

\textbf{Limitations} $\method$ faces certain limitations. For example, collecting revisions can be time-consuming, especially for high-dimensional domains like images. Within layout generation works, our work makes certain assumptions, such as the types of available elements. Future work in enabling for incorporating new assets that designers may find relevant may address this.

\newpage
\textbf{Broader Impact}
Aligning large, generative models, including text-to-layout models, with human values is may help create safe, trustworthy models. In this paper, we hope that by finetuning a pretrained text-to-layout model on a revision dataset, we can improve the quality and possibly safety of generated layouts. Currently, we intend for these text-to-layout models to be used as starting layouts for designers, to help them accelerate the design process of new layouts, rather than used directly and presented to a large audience. Thus, we hope that any dangerous outputs will, firstly, be reduced through our RLHF alignment process, and that secondly, designers will be able to fix or discard any suboptimal layouts during their design process. 

\bibliography{example_paper}
\bibliographystyle{icml2024}

\newpage
\appendix
\onecolumn
\section{Appendix}
\subsection{Hyperparameters}
\subsubsection{PLay Pretraining}
We train a text-conditioned PLay model. We remove the guideline condition and replace it with a text condition, which uses text embedding features from a BERT model with 12 layers, 12 attention heads, and hidden size 768. We inject text conditions through element-wise condition on pooled text embeddings and cross attention with the full text embedding.

The rest of the hyperameters that we used are equivalent to those in \citet{cheng2023play}. We train the model on 8 Google Cloud TPU v4 cores for 40,000 steps with batch size 1024.

\subsubsection{Reward Model Training}
We include the hyperparameters for training in Table~\ref{tab:appendix_rew_model}. 

\begin{table}[h]
\centering
\begin{tabular}{llll}
\toprule
                      & \multicolumn{2}{c}{\textbf{Reward Model Pretraining}}                    & \textbf{Finetuning}         \\
\textbf{Method}       & \textbf{CLAY Pretrain Steps} & \textbf{$\mathcal{D}_{\text{human}}$ Train Steps} & \textbf{Optim. Steps} \\
\midrule
Supervised Finetuning & x                               & x                                      & 1,000                      \\
RARE Keystroke                 & 2,000                           & 200                                    & 1,000                         \\
RARE Chamfer        & 2,000                          & 400                                  & 1,00   \\
Preference     & 2,000                           & 1,00                                    & 800                         \\
\bottomrule
\end{tabular}
\caption{Reward Model Training Hyperparameters.}
\label{tab:appendix_rew_model}
\end{table}

$\method$ and the Preference Reward Model have the same architecture as the denoising diffusion model used in PLay, with the exception that there is no time embedding, and there is an additional MLP layer that reshapes the output features and projects it to a scalar prediction. For the Chamfer Reward Model, we reduce the number of layers to 2, number of heads to 4, and key, query and value dimensions to 256 to prevent overfitting.

\subsection{RLHF Hyperparameters}
\label{appendix:rlhf_hp}
We train with sample batches of 256. In accordance with DDPO, we compute losses for a single timestep across denoising timesteps together. We set the PPO clip range to 1e-2. We use a batch size of 64 on 8 Google Cloud TPU v4 cores.

\subsection{PLay Color Legend}
\begin{figure}[h!]
    \centering
    \includegraphics[width=0.5\textwidth]{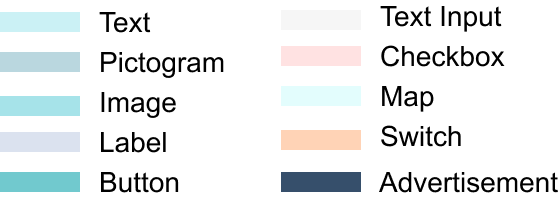}
    \caption{\textbf{Visualization Colors}}
    \label{fig:play_colors}
\end{figure}

We use the same color legend as in \citet{cheng2023play} to visualize the layouts. Colors for popular class elements are rendered in Figure~\ref{fig:play_colors}.

\subsection{Additional Dataset Statistics}
\subsubsection{Procedurally Generated Pretraining Dataset}
\label{appendix:proceduralpretrain}
We procedurally generate synthetic layout pairs to pretrain our reward model. We sample an fully revised layout, and manually perturb elements to mimic a possible intermediate layout that is inferior in quality to the final layout.

To generate a synthetic intermediate layout, we randomly resize them to between $0.5$ to $2.0$ times of their original widths and heights, and we randomly move the elements uniformly between one width and height lower than its original position, and one width and height higher than its original position, bounded by the edges of the layouts. We randomly drop and add generated elements up to $1.5$ times the original numbers of elements in the original sequences. 

Then, we can use these pairs directly as \textbf{Preference} pairs; calculate the \textbf{Chamfer Distance} between the synthetic intermediate layout and true final layout; and generate \textbf{Keystroke times} between pairs by heuristically defining revision distances. 

For the Keystroke times, we heuristically calculate Keystroke times based on our analysis of the edited dataset. For instance, in our revision logs, removing an element is logged as one step, and adding an element may take multiple steps (create element log; modify size log; move object log). Thus, for each dropped element, we assign a cost of $1$ time step to it, and $2$ time step for each revised element, and $3$ time step for each added element.  We also normalize procedurally generated time durations and dataset time durations to minimize discrepancies.

\subsubsection{CLAY}
\begin{figure}[h!]
    \centering
    \includegraphics[width=0.5\textwidth]{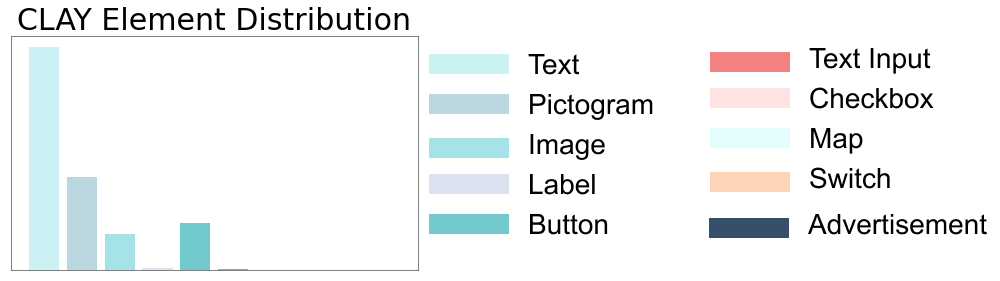}
    \caption{\textbf{CLAY Dataset Element Distribution}}
    \label{fig:clay_distribution}
\end{figure}

PLay is trained on CLAY, a large-scale dataset of 59,555 mobile screen layouts. The average number of elements for the original version of CLAY is 19.6, and it contains 24 element classes, including compound class types such as list items and container. To reduce the complexity of editing sequences and increase the number of revised designs we can collect given the limited time and budget, we select 10 classes to simplify the layouts, and the updated mean number of elements per layout is 11.4 (standard deviation 9.0). The distribution of element classes we train on is shown in Figure~\ref{fig:clay_distribution}.

\subsubsection{Revision Dataset}
Across the revision dataset, designers make on average 889.3 (standard deviation 612.5) edits. Because the logs are extremely verbose, we condense the sequence of revisions to be every 10 logged steps. Excluding extraneous logs that are not reflected in the PLay layouts (e.g. color or font of an element does not affect the vectorized layout), 89.7\% of edits involve rearranging elements and 10.3\% involve resizing elements. This is reasonable, as precisely aligning elements and organizing the layout is a more tedious and common part of revision than resizing elements.

\begin{figure}[h!]
    \centering
    \includegraphics[width=0.5\textwidth]{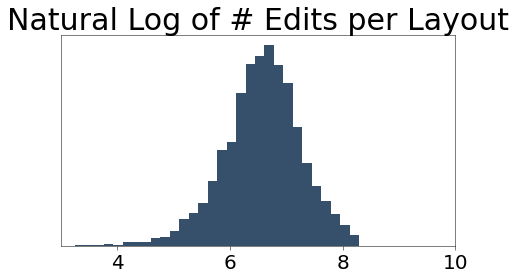}
    \caption{\textbf{Distribution of Number of Edits per Layout}}
    \label{fig:lognum_edits}
\end{figure}

In addition, we provide a histogram of the natural log of the number of edits in Figure~\ref{fig:lognum_edits}.

\begin{figure}
\includegraphics[width=\linewidth]{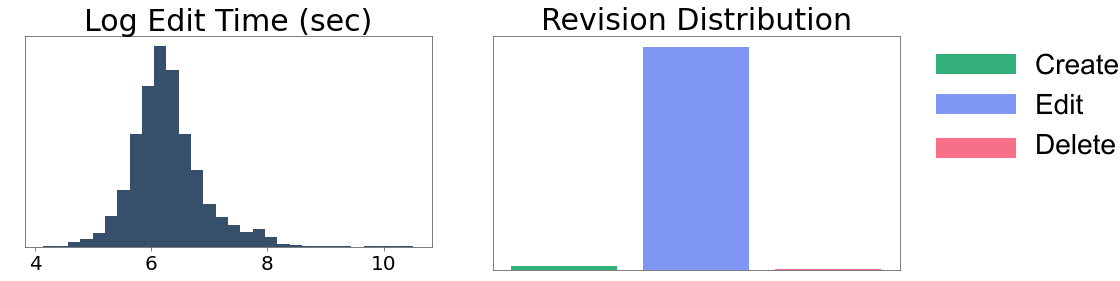}
\label{fig:sfig1}
\caption{\textbf{Revision Statistics} We plot the distribution of natural log of the edit times (left) and the distribution of types of revisions (right). The median time is 503.4 seconds for a designer to complete a full revision. The overall distribution is skewed by a couple of outliers that took an extremely long time, and when taking the natural log of the edit time, it resembles a normal distribution.}
\label{fig:dataset_stats}
\end{figure}

We attach Revision Statistics in Figure~\ref{fig:dataset_stats}.

\subsection{Additional Samples}
\begin{figure}
    \centering
    \includegraphics[width=0.8\linewidth]{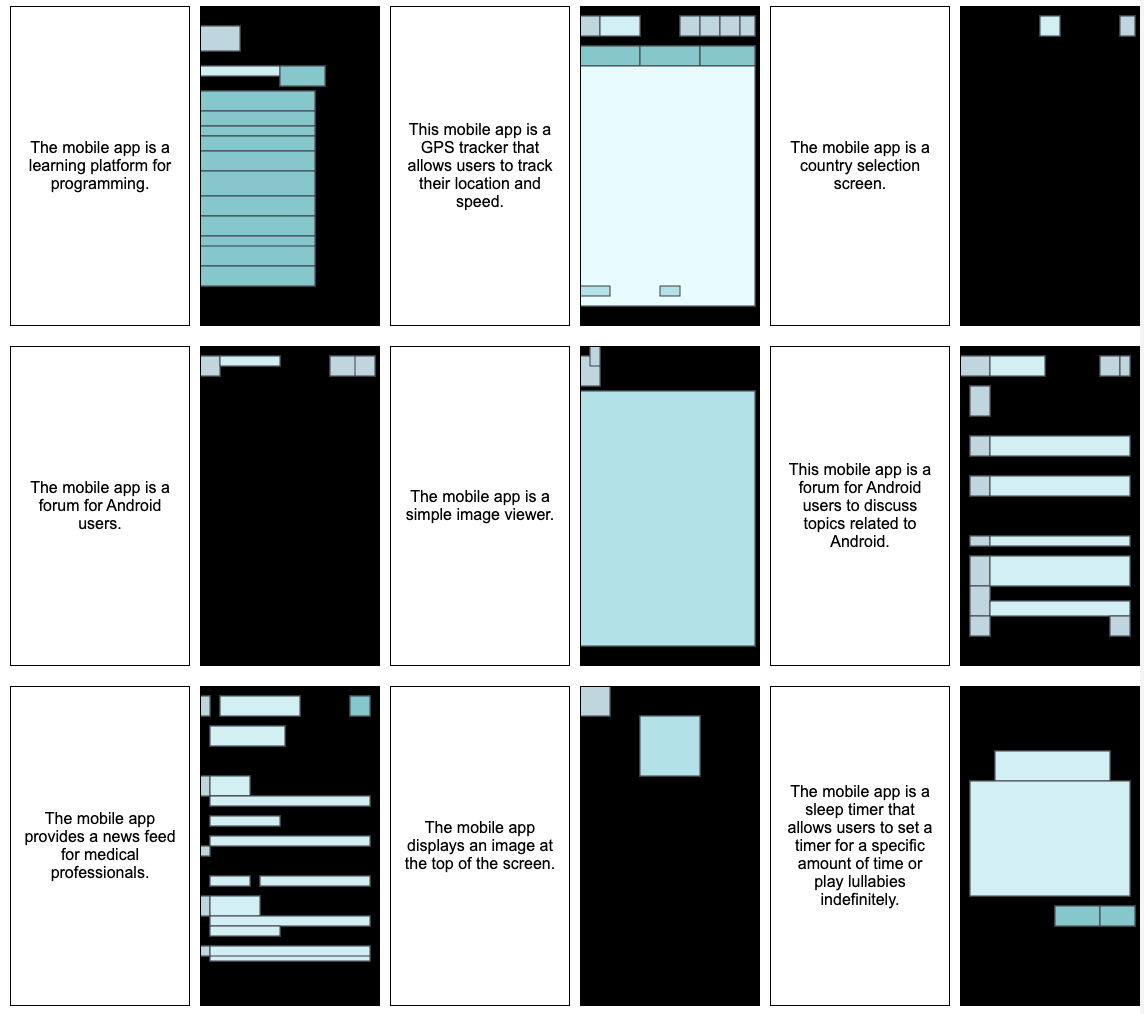}
    \caption{Non-cherrypicked samples from RLHF w/ $\method$ Keystroke}.
    \label{fig:RARE_samples}
\end{figure}

\begin{figure}
    \centering
    \includegraphics[width=0.8\linewidth]{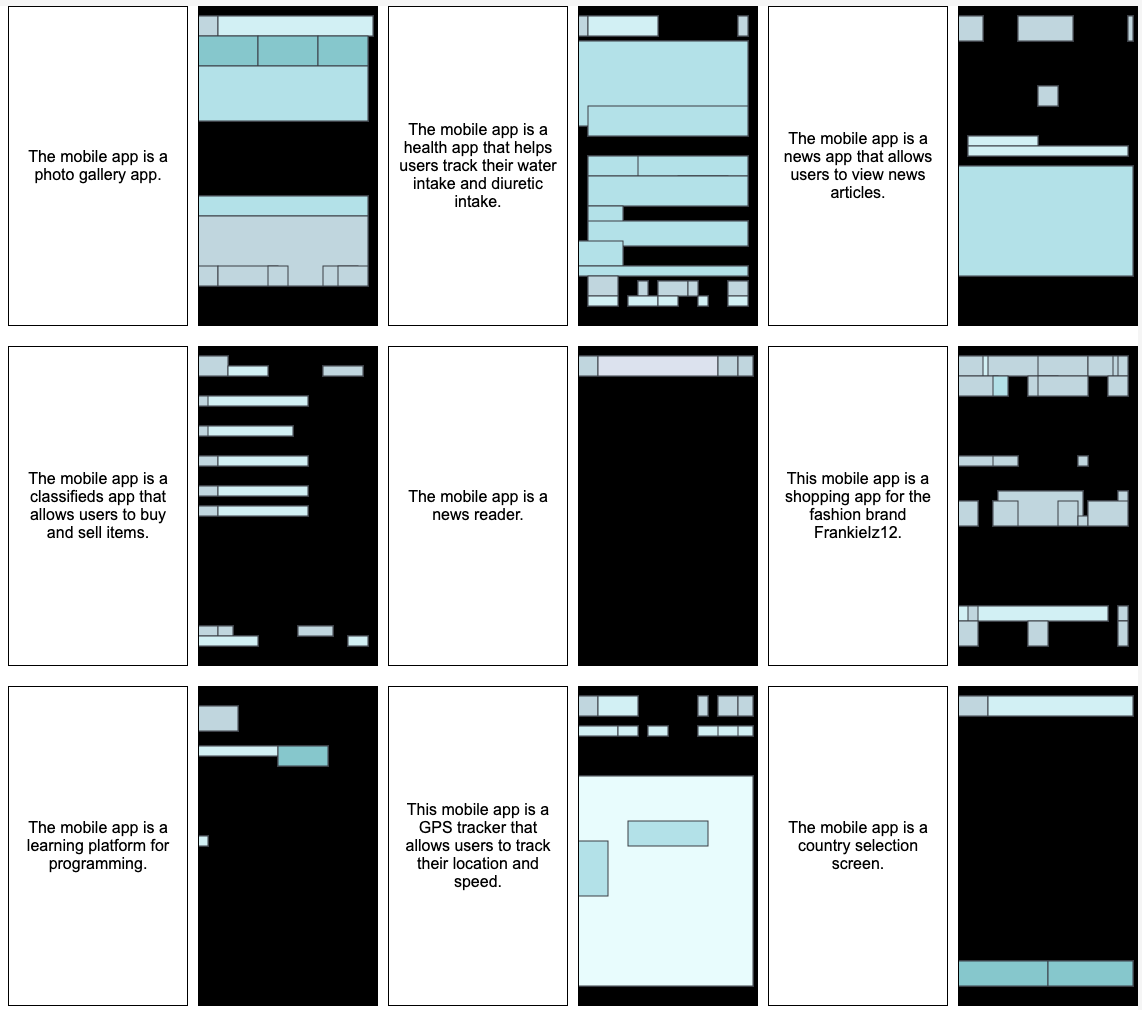}
    \caption{Non-cherrypicked samples from RLHF w/ a Preference-based reward model.}
    \label{fig:preference_samples}
\end{figure}

\begin{figure}
    \centering
    \includegraphics[width=0.8\linewidth]{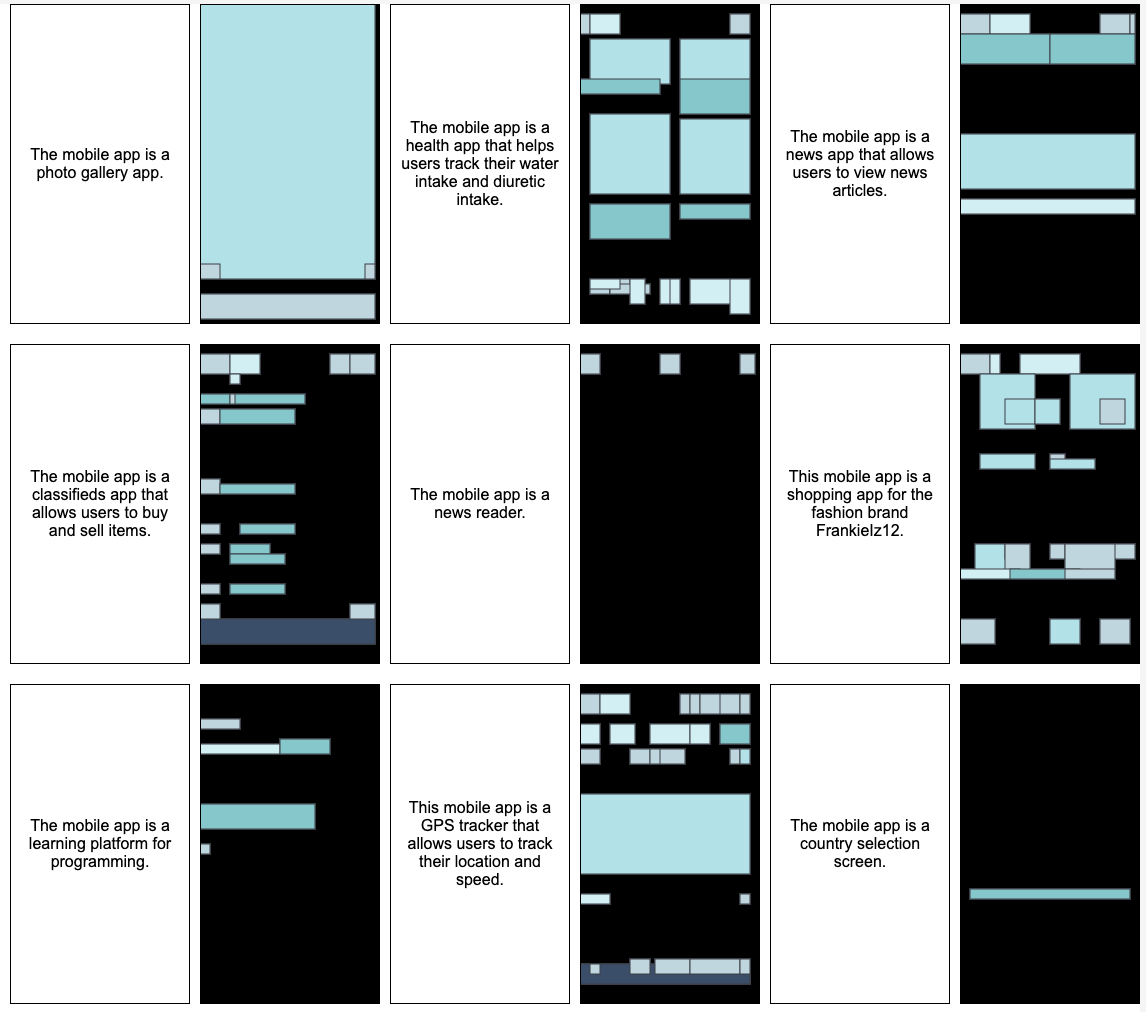}
    \caption{Non-cherrypicked samples from RLHF w/ a $\method$ Chamfer Distance reward model.}
    \label{fig:chamfer_samples}
\end{figure}

\begin{figure}
    \centering
    \includegraphics[width=0.8\linewidth]{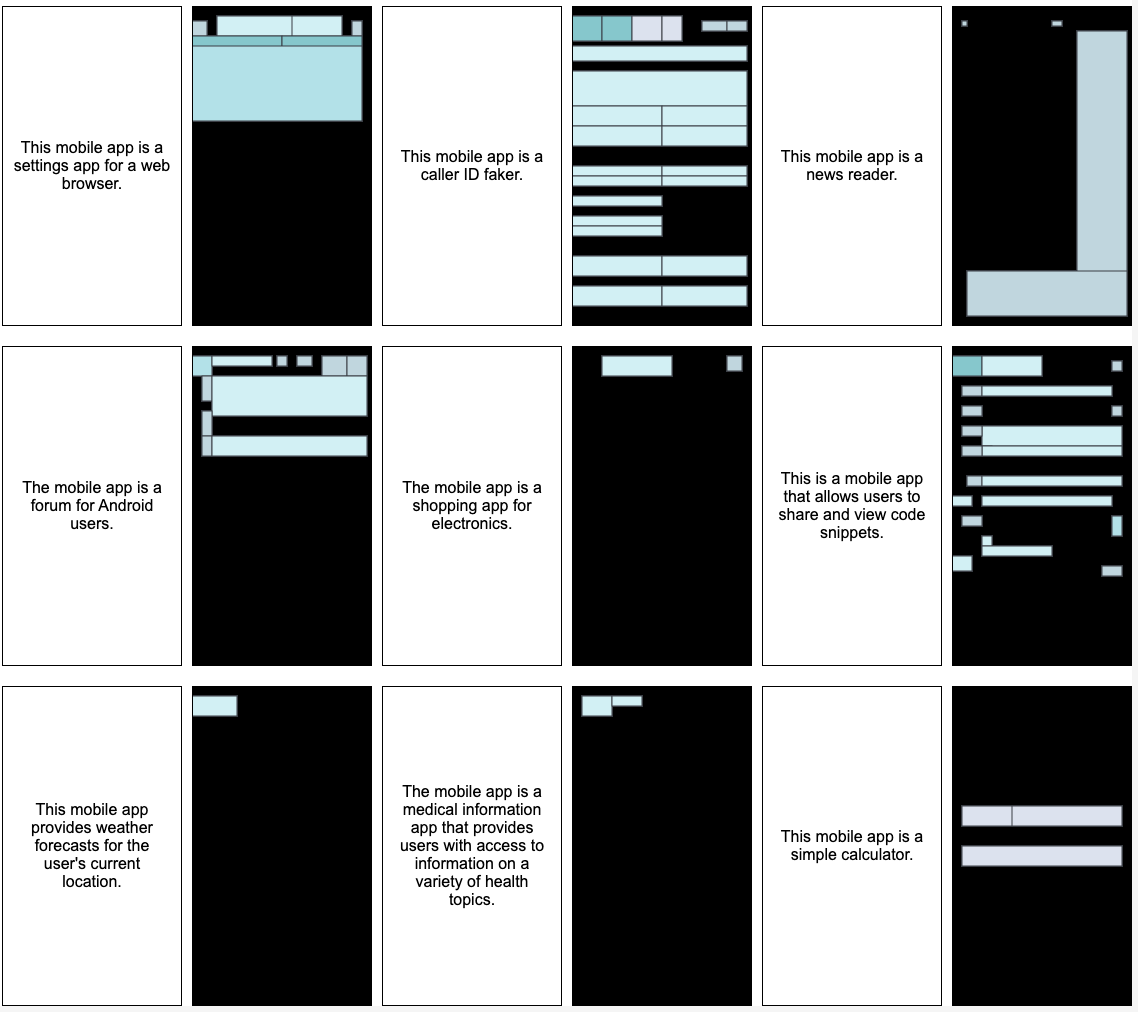}
    \caption{Non-cherrypicked samples from the Supervised Finetuning model.}
    \label{fig:sft_samples}
\end{figure}

\end{document}